\documentclass[aps, prr, twocolumn,showkeys,showpacs,amsmath,amssymb,superscriptaddress, longbibliography]{revtex4-1}

\usepackage{lineno,hyperref}
\modulolinenumbers[5]

\usepackage{makeidx}
\usepackage{graphics} 
\usepackage{color}
\usepackage{listings}
\usepackage{bm}
\usepackage{url}
\usepackage{booktabs}
\usepackage{amsmath}
\usepackage{paralist}
\usepackage{amssymb}
\usepackage{amsthm}
\usepackage{subcaption}
\usepackage{algorithm}
\usepackage{algorithmic}
\usepackage{cancel}
\usepackage{makecell} 

\usepackage{svg}
\usepackage{booktabs}

\definecolor{orange}{rgb}{1,0.5,0}
\definecolor{forestgreen}{rgb}{0.13, 0.55, 0.13}
\definecolor{bittersweet}{rgb}{1.0, 0.44, 0.37}
\definecolor{chartreuse}{rgb}{0.87, 1.0, 0.0}
\definecolor{darkorchid}{rgb}{0.6, 0.2, 0.8}
\definecolor{amethyst}{rgb}{0.6, 0.4, 0.8}

\begin{document}
  
\title{A kinetic-based regularization method for data science applications}

\author{Abhisek Ganguly}
\affiliation{
 Engineering Mechanics Unit, Jawaharlal Nehru Centre for Advanced Scientific Research,\\
 Jakkur, Bengaluru 560064,  India\\
}
\author{Alessandro Gabbana } 

\affiliation{CCS-2 Computational Physics and Methods, Los Alamos National Laboratory, Los Alamos, 87545 New Mexico, USA}
\affiliation{Center for Nonlinear Studies (CNLS), Los Alamos National Laboratory, Los Alamos, 87545 New Mexico, USA}

\author{Vybhav Rao}
\affiliation{
 AI and Robotics Technology Park, Indian Institute of Science, Bengaluru 560012,  India\\
}
\author{Sauro Succi  } 
\affiliation{
  Fondazione Instituto Italiano di Tecnologia,\\
Center for Life Nano-Neuroscience at la Sapienza, Viale Regina Elena 291, 00161 Roma, Italy\\
}
\author{Santosh Ansumali}
\affiliation{
 Engineering Mechanics Unit, Jawaharlal Nehru Centre for Advanced Scientific Research,\\
Jakkur, Bengaluru 560064,  India\\
}

\begin{abstract}
We propose a physics-based regularization technique for function learning, inspired by statistical mechanics. 
By drawing an analogy between optimizing the parameters of an interpolator and minimizing the energy of a system, 
we introduce corrections that impose constraints on the lower-order moments of the data distribution. 
This minimizes the discrepancy between the discrete and continuum representations of the data, 
in turn allowing to access more favorable energy landscapes, thus improving the accuracy of the interpolator. 
Our approach improves performance in both interpolation and regression tasks, even in high-dimensional spaces. 
Unlike traditional methods, it does not require empirical parameter tuning, 
making it particularly effective for handling noisy data.
We also show that thanks to its local nature, the method offers computational and memory efficiency advantages over Radial Basis Function interpolators, especially for large datasets. 

\end{abstract}

\maketitle

\section{Introduction}\label{sec:introduction}

Big data science and accompanying machine learning (ML) techniques have profoundly impacted 
both scientific and societal domains, transforming our ability to collect, store, and navigate 
unprecedented amounts of data~\cite{jarrah2015,lecun-nature-2015,heureux2017}. 
This revolution has been driven by a synergy of key advancements: the ever increasing availability 
of extensive datasets generated e.g. by simulations, experiments, and observational systems; 
the rapid increase in computational power enabled by modern parallel architectures, such as GPUs, 
which are ideally suited for the matrix-vector operations central to ML algorithms; 
the widespread availability of open-source frameworks, which has significantly streamlined 
the development and deployment of ML models.
The transformative power introduced by these techniques has recently been acknowledged through 
the 2024 Nobel Prizes in Physics and Chemistry, underscoring the tremendous impact of ML 
in scientific discovery and technological progress~\cite{hopfield1982,jumper2021}.

Despite the remarkable successes and rapid progress of ML, several limitations remain.
To start with, a key challenge is the generalizability of ML models beyond their training data. 
While ML models perform well on familiar data, they often struggle with unseen scenarios.
This issue arises from several factors. For example, the sheer number of free parameters 
of modern ML models, often reaching hundreds of billions in recent large language model (LLM) 
applications~\cite{touvron2023llama}, can easily lead to overfitting of the training data.

Moreover, while ML excels in uncovering complex correlations in data, in general does not capture 
causal relationships or physical laws.
This limitation impacts the robustness and interpretability of models, particularly in scientific 
applications where understanding underlying physical principles is crucial. 
These problems are well-recognized by the ML community, which has increasingly focused on the 
development of explainable AI and causal inference methods~\cite{Pearl2000,scholkopf2021toward}. 

A common approach for addressing some of these issues is regularization, which can help mitigate 
overfitting and enhance generalization. 
Regularization techniques can be divided into two broad categories: non-physics-based and 
physics-based approaches. 
Non-physics-based regularization techniques, such as L2 (ridge regression)~\cite{hoerl-1970-ridge} 
and L1 (lasso)~\cite{tibshirani-1996-lasso} regularization, have long been standard tools in data science, 
and aim at reducing the complexity of the model by adding penalties to the model parameters. These techniques 
involve tuning of hyperparameters through cross-validation or trial and error, which can be costly and time-consuming. A more efficient and systematic approach is Maximum Likelihood Estimation (MLE), which chooses hyperparameters by maximizing the marginal likelihood~\cite{rasmussen2006gaussian}.

In contrast, physics-based regularization incorporates domain-specific knowledge into the learning 
process, ensuring that the learned model complies to fundamental physical laws, such as conservation 
of mass, momentum, energy~\cite{karpatne-2017-constrains,raissi-2019-pinn,bronstein-2021}.
This is precisely the goal of what is known as Physics Informed Machine Learning (PIML)~\cite{willard-2020-piml,hao-2022-piml}, 
which aims at incorporating underlying physical structures or constraints in ML models. 

PIML methods help bridging the gap between data-driven learning and the underlying physical 
principles, enhancing the model's robustness and ability to generalize beyond the training set. 
For example, enforcing constraints on preservation of symmetry or conservation laws, allows the 
model to remain physically consistent, even when making predictions for scenarios not included 
in the training set. Often, the kernels used in these models are also inspired by physics. In Higher Dimensional Model Representation using Gaussian Process Regression (HDMR-GPR)~\cite{Rabitz1999hdmr}, for instance, the kernel is constructed as a sum of individual Gaussian functions corresponding to increasing orders of interaction, an approach inspired by many-body interactions in quantum and statistical mechanics.
However, hard-wiring physical constraints into ML algorithms can be a challenging task~\cite{beucler-prl-2021}. 
Often, soft constraints are employed in the form of penalty terms within 
the loss function to guide optimization. While this approach is relatively simple, 
the downside is that it is not guaranteed that the constraints will be strictly enforced in all cases.

In this context, we introduce a novel regularization approach, 
inspired by kinetic theory, which can be applied to the `learning'
of in principle any complex function in high-dimensional spaces,
without prior knowledge of the underlying system.
\begin{figure}[htb]
  \centering
  \includegraphics[width=0.99\columnwidth]{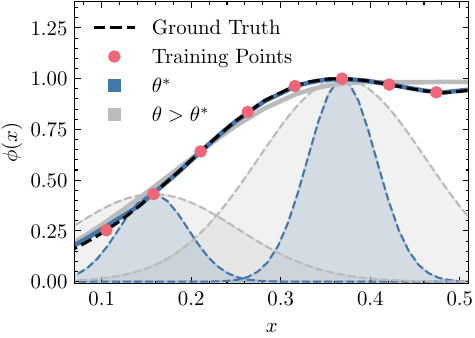}
  \caption{Notion of data temperature as proxy to data-correlations: The interpolated curves are generated by the means of overlapping Gaussians associated with the training points, the temperature $\theta$ and centers being the parameters. An optimum set of these two parameters can be found resulting in the best fit. The figure shows two cases, comparing the optimal temperature $\theta^{*}$, with a higher temperature which under-fits the dataset.
          }\label{fig:dataTemp}
\end{figure}

The learning process offers an interpretation in terms of statistical 
mechanics, where the goal is to define a mapping between 
a system's microstate (the input variables) and its macrostate (corresponding outputs). 
The objective is hence to uncover the underlying generative structure of the data, 
which, in statistical terms, translates to identifying the probability distribution governing the system.

Building on this idea, and drawing from lessons learned in the development of 
discrete kinetic numerical solvers~\cite{succibook2018}, we demonstrate the
benefit of constraining the lower order moments of the distribution, such to
minimize the discrepancy between their continuum and discrete counterpart.

The procedure can be seen as a powerful and physically grounded 
regularization technique, beneficial for both interpolation and regression tasks. A pictorial representation of the procedure to interpolate a curve is given in Fig.~\ref{fig:dataTemp} as a method overview.

The remainder of this paper is organized as follows. In Section~\ref{sec:theory} we introduce 
a class of \textit{thermal} interpolators, and provide details on the regularization procedure.
In Section~\ref{sec:numerics} we benchmark the method in both interpolation and regression tasks, 
highlighting the gains provided by the regularization procedure, also comparing with other 
standard methods.
Finally, Section~\ref{sec:conclusion} concludes the paper, with discussion on the relative 
advantages and limitations of this regularization method and also potential future works.

\section{Theory and Methodology}\label{sec:theory}

The process of understanding the data generative pattern in a system can be formalized in 
statistical terms as identifying the underlying probability distribution that governs 
the relationship between inputs and outputs~\cite{BishopMmodel}. 
For a set of $D$-dimensional input data points
${\bm X} = \{{\bm x}_1, {\bm x}_2, \dots, {\bm x}_N\}$ 
and their corresponding output states $\Phi = \{\phi_1, \phi_2, \dots, \phi_N\}$, 
this relationship can be expressed as
\begin{equation}\label{eq:phi-basic}
  \phi({\bm x}) = \sum_{i=1}^N p({\bm x},{\bm x}_i)\phi_i ,
\end{equation}
Here, $\phi({\bm x})$ represents the predicted value at a new input ${\bm x}$, 
while the discrete kernel function $p$ quantifies how much each data point ${\bm x}_i$ 
contributes to the prediction, in turn bridging the discrete set of training points to a continuum. 
Although normalizing the kernel function is not mandatory, it is often beneficial since it allows 
$p$ to be interpreted as a discrete probability distribution, defined as
\begin{equation}
    P(\bm{x},\bm{x_i}) = \frac{p(\bm{x},\bm{x_i})}{\sum_{i=1}^N p(\bm{x},\bm{x_i})}
\end{equation} 

Different approaches, such as radial basis function (RBF)~\cite{OrrRadial,orr1996introduction} 
and Neural Networks (NN)~\cite{Goodfellow-et-al-2016} differ in their choice of the probability distribution function 
and the specific manipulations applied to the local functional.  
A more general version of Eq.~\ref{eq:phi-basic} can be expressed as
\begin{equation}\label{eq:phi-general}
  \phi({\bm x}) = \sum_{i=1}^N P({\bm x},{\bm x}_i) \, \psi( \phi_i ({\bm x}_i) ),
\end{equation}
where the transformation function $\psi$ is applied to the output values $\phi_i ({\bm x}_i)$
at the training points, allowing the model to capture more complex, non-linear relationships 
between the input and output spaces. In the context of NN the function $\psi$ is commonly 
known as activation function, with typical examples being sigmoid, softmax, ReLU etc.
\begin{algorithm}[H]
  \caption{Interpolator with kinetic-based regularization}\label{alg:summary}
  \begin{algorithmic}[1]
  \REQUIRE Input data $\{\bm{x}_i, y_i\}_{i=1}^N$, distribution function $P(\|\tilde{\bm{x}} - \bm{x}_i\|; \theta)$ for a given $\theta$, local polynomial basis $\psi(\bm{x})$
  \ENSURE Approximation $\hat{\phi}(\bm{x})$ at unknown points

  \STATE \textbf{Step 1: Local Linear Fit:}\\
   Calculate the Lagrange Multiplier $\delta \bm{x}$ using the iterative solution of Eq.~\eqref{eq:LM_1D} or Eq.~\eqref{eq:LM_general} depending on $D$ to satisfy the first moment condition:
    \[
  \sum_i \bm{x}_i P(||\bm{x}+\delta \bm{x} - \bm{x}_i||, \theta) = \hat{\bm{x}},
  \]
  \[
  \tilde{\bm{x}} = \bm{x} \, + \, \delta \bm{x}.
  \]
    \STATE \textbf{Step 2: Normalization:}\\
   Ensure zeroth moment satisfaction by normalizing the chosen kernel function $p$ and creating a discrete probability distribution function $P$:
    \[
   P(||\tilde{\bm{x}} - \bm{x}_i||, \theta) = \frac{p(||\tilde{\bm{x}} - \bm{x}_i||, \theta)}{\sum_j p(||\tilde{\bm{x}} - \bm{x}_j||, \theta)}.
  \]
 
  \STATE \textbf{Step 3: Locally Quadratic Fit}\\
  Satisfy the second moment by first predicting values on the known set of input data $\bm{x}_i$. 
  \STATE \textbf{Step 4: Prediction on Training Set}\\
  Compute:
  \[
  \hat{\phi}(\bm{x}_i) = \sum_j \phi(\bm{x}_j) P(||\bm{x}_i + \delta \bm{x}_i - \bm{x}_j||; \theta).
  \]
  \STATE \textbf{Step 5: Self-Correction}\\
  Calculate the polynomial basis $\psi(\bm{x})$ at input points $\bm{x}_i$:
  \[
  \psi(\bm{x}_i) = 2\phi(\bm{x}_i)-\hat{\phi}(\bm{x}_i).
  \]
  \STATE\textbf{Step 6: Prediction}\\
  Adjust predictions using the error correction at unknown points $\bm{x}$ and the modified distribution function $P$:
  \[
  \hat{\phi}(\bm{x}) = \sum_i \psi(\bm{x}_i) P(||\tilde{\bm{x}} - \bm{x}_i||;\theta).
  \]

  \end{algorithmic}
\end{algorithm}

For RBF, instead, Eq.~\ref{eq:phi-general} reduces to
\begin{equation}\label{eq:phi-rbf}
  \phi({\bm x}) = \sum_{i=1}^N P( \| \bm{x} - \bm{x}_i \| ) \psi({\bm x}_i)  ,
\end{equation}
where $\psi({\bm x}_i)$ are the (unknown) weights, and $P$ now depends explicitly on the Euclidean distance 
between the input point ${\bm x}$ and each training point $\bm{x}_i$.
In order to compute the weights, one needs to impose the condition $\phi( {\bm{x}_i} ) = \phi_i$,
leading to a linear system of equations where for each training point 
\begin{equation}\label{eq:rbf-cond}
  \phi_i = \sum_{j=1}^N P( \| \bm{x}_i - \bm{x}_j \| ) \, \psi({\bm x}_j)  .
\end{equation}
It follows that RBF requires inverting/factorizing a $N \times N$ matrix, associated 
in the worst case scenario (i.e. employing direct methods) with a computational cost $\mathcal{O}(N^3)$, 
which is prohibitively expensive for very large datasets. 
However, iterative solvers -- such as the Conjugate Gradient method with appropriate preconditioning -- 
can reduce the cost significantly (each iteration typically incurs a cost of roughly $\mathcal{O}(N^2)$, 
and if the number of iterations remains modest, the overall cost becomes substantially lower than 
$\mathcal{O}(N^3)$.

In what follows we will introduce a new method, which albeit similar in spirit to RBF, 
takes a very different pathway for the calculation of $\psi$. 
We start from Eq.~\ref{eq:phi-rbf}, and assume, without any loss of generalization,  
$P({\bm x},{\bm x}_i) \equiv P(\|{\pmb \xi}_i\|,\theta)$, with $ {\pmb \xi}_i = \bm{x} -  \bm{x}_i   $,
and with the distribution shape dictated by the single free parameter $\theta$.
For simplicity, we hereafter restrict the discussion and the numerical analysis
to the Gaussian distribution: $ P(\|{\pmb \xi}_i\|,\theta) = 
\left( 2\pi \theta\right)^{- D/2}\, \exp{\left(-\|{\pmb \xi}_i\|^2 /(2\theta)\right)}$.

\subsection{Regularization procedure: Constraining the low-order moments of the distribution}

Instead of relying on the global condition stemming from Eq.~\ref{eq:rbf-cond}, we compute 
$\psi$ by imposing conditions on the moments of the distribution function. 
Let us start assuming that $\phi$ is a locally linear function in the neighborhood of $\bm{x}$,
hence $\phi({\bm x}) = a + \bm{b} \cdot \bm{x}$.
In order for the interpolator to satisfy the locality of $\phi$ we shall require:
\begin{align}
  \sum_i          P( \| \pmb \xi_i \|,\theta) &= 1,      \label{eq:zero_moment} \\
  \sum_i \bm{x_i} P( \| \pmb \xi_i \|,\theta) &= \bm{x}. \label{eq:first_moment}
\end{align}
Conditions on higher order moments of the distribution can be derived assuming $\phi$ to be 
locally quadratic, cubic, etc.

We shall remark that while the conditions in Eq.~\ref{eq:zero_moment} and Eq.~\ref{eq:first_moment} 
hold true in the continuum, in general they might not be satisfied in the discrete. 
The problem of matching discrete and continuous moments was resolved using discrete 
kinetic theory for the development of Lattice Boltzmann methods~\cite{succibook2018} 
e.g. via the method of entropy maximization~\cite{karlin1998, ansumali2003}. 

We now detail the procedure for ensuring that the moments of the discrete and continuum distribution
are matched up to second order.

\subsubsection{Zeroth moment}\label{sec:zeroth_moment}
The condition given by Eq.~\ref{eq:zero_moment} implies a specific normalization of the 
discrete probability distribution. For the Gaussian function we have 
\begin{equation}\label{eq:normalized-gauss}
  P(\| \pmb \xi_i \|, \theta) 
  = 
  \frac{ \exp(- \| \pmb \xi_i \|^2 / (2\theta))}{\sum_j \exp( - \| \pmb \xi_j \|^2 / (2\theta))}.
\end{equation} 

\subsubsection{First moment}\label{sec:first_moment}
 
We now take into consideration the condition on the first moment of the distribution 
from Eq.~\ref{eq:first_moment}.
Observe that the zero temperature limit ($\theta \rightarrow 0$) of the Gaussian distribution,
i.e. $P({\bm x},{\bm x}_i)=\delta({\bm x}-{\bm x}_i) / \sum_j \delta({\bm x}-{\bm x}_j)$ satisfies
the condition, although only within the training data set ($\bm{x} \in \bm{X}$).

In order to ensure that Eq.~\ref{eq:first_moment} holds in general, we start by defining a correction
to the generic displacement vector, which we denote as $\tilde{\pmb \xi}_i = \bm{x}_i - \tilde{\bm{x}}$,
where this time $\bm{x} \notin \bm{X}$. We then use $\tilde{\pmb \xi}_i$ to define a modified 
discrete probability distribution, yielding the following condition on the first moment:
\begin{equation}\label{eq:rootproblem}
  \sum_i (\bm{x} - \bm{x_i}) P( \| \tilde{\pmb \xi}_i \|, \theta) = 0.
\end{equation}
The above is a transcendental equation which needs to be solved for $\tilde{\bm{x}}$.
This can be achieved using a iterative method. We leave full details on the solution 
of Eq.~\ref{eq:rootproblem} to Appendix~\ref{sec:a1}, stating here the expression for the
$k^{th}$ iteration for the 1D case as an example:
\begin{gather}\label{eq:LM_1D}
  \tilde{x}^{(k+1)} 
  = 
  \tilde{x}^{(k)} 
  + \theta \frac{ %
                  \sum_i (x - x_i) P(\| \tilde{x}^{(k)} - x_i \|, \theta)
                }
                { %
                  \sum_i\left((\tilde{x}^{(k)}-x_i )(x-x_i) P( \| \tilde{x}^{(k)} - x_i \|, \theta)\right)
                } .
\end{gather}
Starting from e.g. $\tilde{x}^{(0)} = x$, the above is iterated until Eq.~\ref{eq:rootproblem}
is satisfied within the prescribed accuracy of the root-finding solver.
We shall stress that as the number of spatial dimensions $D$ increases, the dimension of
Eq.~\ref{eq:rootproblem} increase accordingly, with the system size being, for each training point, 
$D \times D$, which is a strong improvement over the $N \times N$ requirement of RBF.

\subsubsection {Second moment}\label{sec:second_moment}
We consider once again Eq.~\ref{eq:phi-general}, having applied to $P$ the corrections required 
to match zeroth and first moments, and with $\psi(\phi({\bm x}_i)) = \phi({\bm x}_i)$:
\begin{equation}\label{eq:phi-tilda}
  \hat{\phi}({\bm x})
  = 
  \sum_i \phi({\bm x}_i) P(\|\tilde{\pmb \xi}_i \|,\theta) .
\end{equation}
We now assume $\phi$ to be a locally quadratic function in the neighborhood of $\bm{x}$:
\begin{equation}\label{eq:local-quadratic}
  \phi({\bm x}) = a + \bm{b} \cdot \bm{x} + \bm{x}^T \bm{C} \bm{x}
\end{equation}
with $\bm{C} \in \mathcal{R}^{D \times D}$ a symmetric matrix.

Inserting Eq.~\ref{eq:local-quadratic} into Eq.~\ref{eq:phi-tilda} leads to the 
following approximation:
\begin{equation}\label{eq:first-approx}
  \phi^{(1)}({\bm x})
  =
  \sum_i   \phi({\bm x}_i) P(\|\tilde{\pmb \xi}_i \|,\theta)
  = 
  \phi({\bm x})+{ \text{tr}(\bm{C})\,\Theta}   .
\end{equation}
In the above, we have introduced the variable $\Theta$ to stress that, while $ \Theta \equiv \theta $ in the continuum case,
in the case of discrete moments may in general exhibit a spatial dependence.

In general, it is not possible to use the above expression to get an estimate of the error,
unless we consider points inside the training set.
Indeed, for $\bm{x}_i \in \bm{X}$, it directly follows from Eq.~\ref{eq:first-approx}
that the error committed is
$\epsilon = \phi({\bm x}_i)-\phi^{(1)}({\bm x}_i) = \text{tr}(\bm{C})\,\Theta({\bm x})$.

Next, we can employ $\psi(\phi(\bm{x}_i)) = \phi^{(1)}({\bm x})$ in Eq.~\ref{eq:phi-tilda},
which delivers the following approximation:
\begin{equation} \label{eq:second-approx}
  \phi^{(2)}({\bm x}_i) 
  =
  \phi({\bm x}_i) + 2~{ \text{tr}(\bm{C})\,\Theta({\bm x})} .
\end{equation}
Combining Eq.~\ref{eq:first-approx} and Eq.~\ref{eq:second-approx}, we can construct 
an improved expression of $\psi$:
\begin{equation}
  \phi^{(2)}({\bm x})
  =
  \sum_i  \left(2\, \phi({\bm x}_i)-\phi^{(1)}({{\bm x}_i})\right)\, P(\|\tilde{\pmb \xi}_i \|,\theta). 
\end{equation}
It is important to note that the cancellation of the spurious quadratic term in the above expression is exact only for points in the training set (and provided that $\Theta$ has no spatial dependence). Nevertheless, a partial improvement can still be expected for predictions on unseen points, as we will show in the coming sections (see also Appendix~\ref{sec:a2} for more details).

We conclude this section presenting in Algorithm~\ref{alg:summary} a summary of all the steps 
required to implement the interpolator and the different corrections based on
the moments of the distribution.

\subsection{Tuning the temperature parameter} \label{sec:theta}
In the previous section we did not address how $\theta$, the only free parameter of our method,
should be selected. %
In analogy with statistical physics, where temperature governs the balance between local 
and global fluctuations, the parameter $\theta$ controls the interaction range between points. 
A meaningful choice of $\theta$ should scale with the square of a characteristic length $d_{\rm typ}$, 
which represents the typical spacing between data points. %
In the low-temperature limit ($\theta \rightarrow 0$), the Gaussian kernel reduces to 
a Dirac delta function. As a result, each point is effectively isolated, 
and interactions with neighboring points vanish. This leads to a ``cold death'' scenario, 
where the model memorizes individual points (over-fitting) with no meaningful interpolation/generalization
for points outside the input dataset. 
Conversely, in the high-temperature regime ($\theta \gg d^2_{\rm typ}$), the Gaussian kernel flattens, 
making all points exert nearly uniform influence. This results in a ``hot death'', where 
the model collapses into a global function, losing its ability to capture fine-grained structures,
thus leading to under-fitting.
To avoid these extremes, a natural scaling is $\theta = k \cdot d^2_{\rm typ}$ with $k$ a free scaling parameter,
while $d_{\rm typ}$ can be estimated as the median or mean pairwise distance, or the average k-nearest neighbor distance.

Alternatively, one can get a first rough estimate for $\theta$ from the variance of the training dataset:
\begin{equation}\label{eq:theta-guess}
  \theta^{(0)} = \frac{1}{N} \sum_{i=1}^N \left(x_i - \bar{x}\right)^2, 
  \quad 
  \bar{x} = \frac{1}{N} \sum_{i=1}^N x_i.
\end{equation}

Starting from this guess one can determine an optimal value for $\theta$ employing standard search algorithms~\cite{nocedal-1999-book}.

We propose here a simpler approach, which does not involve computation of gradients, inspired by the maximum entropy approach to parameter estimation \cite{Dempster},
in which, starting from Eq.~\ref{eq:theta-guess}, the temperature parameter can be updated iteratively 
to satisfy the continuum relation:
\begin{equation}\label{eq:stdDevTMax-Ent}
  \theta^{(n+1)} 
  = 
  \frac{1}{N} \sum_{i=1}^N \left(x_i - \bar{x}\right)^2 P(||{\pmb \xi}_i ||, \theta^{(n)}) .
\end{equation}
Physically, Eq.~\eqref{eq:stdDevTMax-Ent} indicates that the average discrete energy aligns 
with its continuum equivalent. To stabilize and control the convergence, 
the update can incorporate a relaxation parameter $\alpha$, yielding:
\begin{equation}\label{eq:theta_lst}
  \theta^{(n+1)} 
  = 
  \alpha \left[
               \frac{1}{N} \sum_{i=1}^N \left(x_i - \bar{x}\right)^2 P(||{\pmb \xi}_i ||, \theta^{(n)})
         \right] + (1 - \alpha) \theta^{(n)}.
\end{equation}
Here, $\alpha$ determines the influence of the previous estimate $\theta^{(n)}$ 
on the new estimate $\theta^{(n+1)}$. The process iteratively searches for the value of 
$\theta$ that minimizes a validation error, effectively identifying 
the optimal point in the system's energy landscape.
To implement this procedure, one needs to define:
\begin{itemize}
  \item An error metric $\varepsilon$, in order to establish a quantitative measure of prediction error (e.g., mean squared error, mean absolute error).
  \item A splitting of the input data into a training set of size $N_t$, and a validation set 
        of size $N_v$, with $N=N_t + N_v$.
\end{itemize}

The iterative search concludes when the validation error converges, when the update step falls 
below a predefined tolerance, or the maximum number of iterations is reached. In this latter case, 
the value of $\theta$ corresponding to the minimum observed validation error 
is selected as the optimal parameter.
In Appendix~\ref{sec:a4} we compare this approach with the Maximum Likelihood Estimate approach.

\section{Numerical Results}\label{sec:numerics}
\begin{figure*}[!htb]
    \centering
    \includegraphics[width=0.95\textwidth]{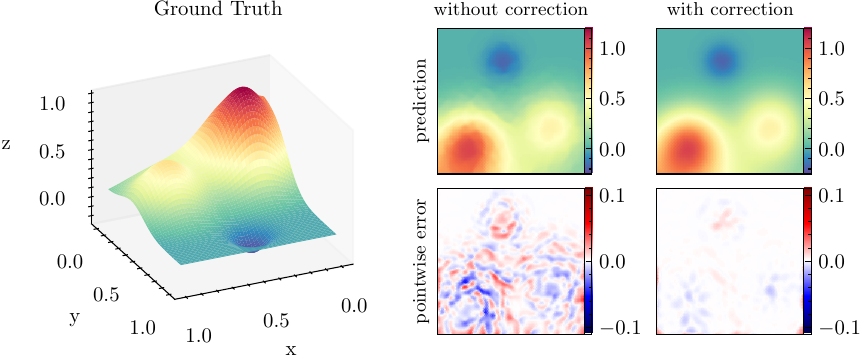}
    \caption{Left: Franke function used as ground truth. 
             Right: Effect of moment corrections on the prediction of the 2D Franke 
             function using 2000 randomly sampled points.  
             The pointwise error plots highlight the benefits of the regularization procedure.
            }\label{fig:franke_noiseless}
\end{figure*}
\begin{figure*}[!htb]
    \centering
    \includegraphics[width=0.85\textwidth]{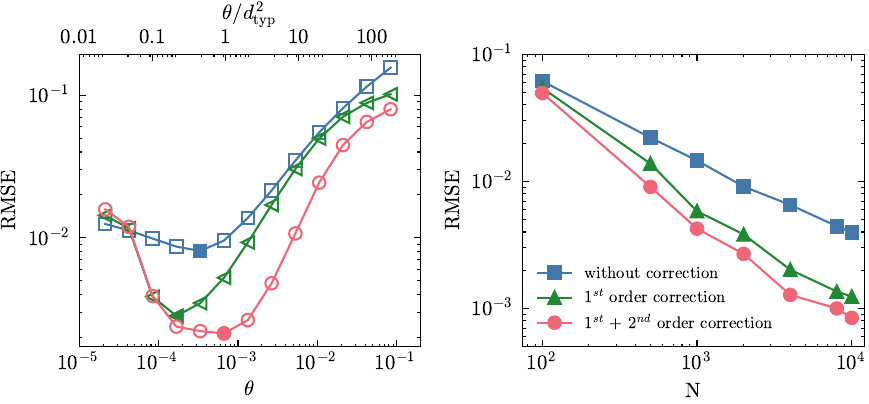}
    \caption{ The figure illustrates the progressive improvements in the prediction of the test function, 
              driven by a decreasing energy minimum in the error landscape as the moment correction order increases.
              Left: RMSE vs. $\theta$ for the 2D Franke function on the validation set $(N_v)$, 
              using $2000$ randomly sampled points $(N = 2000)$ for model training.
              Right: convergence test for the 2D Franke function on a test set of $10000$ uniformly spread points, 
              showing the test RMSE as a function of the sampling size $(N)$.
            }\label{fig:franke_noiseless_analysis}
\end{figure*}
In this section, we evaluate the impact of the moments-based regularization procedure introduced 
in the previous section for both interpolation and regression tasks.
For each test case, we consider a dataset consisting of $N$ points for training the model.
This set is then split into a training dataset of size $N_t$ and a validation set of size $N_v$,
such that $N=N_v + N_t$. These subsets are used to apply the regularization procedure 
and to determine the optimal value of the parameter $\theta$ (see Sec.~\ref{sec:theta}). 
Before training, the data is normalized so that the components of $\bm{x}$ lie within the range $[0,1]$, 
while the base functional values are scaled to $[-1,1]$. 
This normalization provides a consistent reference framework and helps comparison of the error 
committed across different test cases.
Alternatively, one can consider instead a dimensionless representation, scaling quantities by appropriate power of $d_{\rm typ}$ (e.g.  $\bm{\xi}_i/ d_{\rm typ} $, $\theta / d^2_{\rm typ}$, etc).

In what follows we will refer to the following error metrics ($\hat{\phi}(\bm{x})$ being the predicted value):
\begin{gather}
  \langle L_1\rangle         = \;\bigl\langle|\hat{\phi}(\bm{x})-\phi(\bm{x})|\bigr\rangle,\\
  \mathrm{RMSE}   = \;\frac{1}{\max|\phi(\bm{x})|}\,
                      \Bigl\langle\bigl(\hat{\phi}(\bm{x})-\phi(\bm{x})\bigr)^2\Bigr\rangle^{\tfrac{1}{2}},\\
  L_\infty \;=\;\max\,|\hat{\phi}(\bm{x})-\phi(\bm{x})|,
\end{gather}
We also introduce the coefficient of determination $R^2$, defined as:
\begin{equation}
R^2 = 1 - \frac{ \left\langle \bigl(\hat{\phi}(\bm{x}) - \phi(\bm{x})\bigr)^2 \right\rangle }
{ \left\langle \bigl(\phi(\bm{x}) - \langle \phi(\bm{x}) \rangle \bigr)^2 \right\rangle }.
\end{equation}

\subsection{Effect of moment corrections}\label{sec:effect_of_mom_corr}
We start our analysis by evaluating the impact of the regularization procedure on a simple
interpolation task.
We take into consideration the two-dimensional Franke function~\cite{franke-mc-1982}, 
which is commonly employed to test numerical interpolators.
The function (see Fig.~\ref{fig:franke_noiseless}a) is defined as:
\begin{equation}\label{eq:franke}
  \begin{aligned}
    f_{\text{Franke}}(x, y) &= \frac{3}{4} \exp\left(-\frac{(9x - 2)^2}{4} - \frac{(9y - 2)^2}{4}\right) \\
    &\quad + \frac{3}{4} \exp\left(-\frac{(9x + 1)^2}{49} - \frac{(9y + 1)^2}{10}\right) \\
    &\quad + \frac{1}{2} \exp\left(-\frac{(9x - 7)^2}{4} - \frac{(9y - 3)^2}{4}\right) \\
    &\quad - \frac{1}{5} \exp\left(-(9x - 4)^2 - (9y - 7)^2\right).
  \end{aligned}
\end{equation}
We consider a set of $N = 2000$ points, randomly picked in the interval $[0, 1] \times [0, 1]$,
subsequently split in a training and test set using a $80:20$ ratio.
We apply Algorithm~\ref{alg:summary}, and use the test set for establishing the optimal value of
the $\theta$ parameter, or, in other terms, the \textit{temperature} corresponding 
to the local minima in the energy landscape of the predicted system.

In Fig.~\ref{fig:franke_noiseless} we provide a first qualitative comparison between the results 
obtained following this procedure, and the results obtained instead using the plain interpolation method
defined in Eq.~\ref{eq:phi-basic}. For both cases we plot the predicted function values,
as well as the point-wise absolute error with respect to the ground-truth function, highlighting
the benefit of employing the regularization procedure.

Turning on a more quantitative ground, in Fig.~\ref{fig:franke_noiseless_analysis} we investigate
the impact of the different correction terms on the overall accuracy, measured in terms of RMSE.
In the left panel of the figure, we take as an example the case $N = 2000$, and show the 
energy landscape as a function of the temperature parameter $\theta$. It is instructive to observe 
how such landscape gets modified when applying constraints to the moments of the distribution. 
In comparison to the baseline method (blue curve), applying constraints up to the first order of the 
distribution allows to identify an optimal value of $\theta$ for which we can reduce the error 
by a factor $\approx 3$, with a further slight improvement when constraining also the second order moment.
Ultimately, the constraints carve out a nice local minimum in the energy landscape, 
which is the reason why they improve the results.
In the Figure the right-most values on the x-axis are obtained from the ansatz in Eq.~\ref{eq:theta-guess}.
Following the discussion in Sec.~\ref{sec:theta}, we also show a rescaling of the x-axis, given by 
$\theta / d^2_{\rm typ}$, where $d_{\rm typ}$ was taken as the average k-nearest neighbor distance with $k=5$
on the training dataset.
This rescaling highlights the regions corresponding to under-fitting and over-fitting, also 
providing a superior initial guess for identifying the optimal value of $\theta$.

In the right panel of Fig.~\ref{fig:franke_noiseless_analysis} we show the RMSE corresponding to the optimal
value of $\theta$ for training set of increasingly large number of points. The plot shows that the constraints
offer a systematic improvement over the baseline method.

In order to extend the analysis to a general D-dimensional setting,
we next consider the following two-humped camel function 
\begin{multline}\label{eq:camel-ddim}
  f_{\text{camel}}(\mathbf{x}) = \frac{1}{2(k \sqrt{\pi})^{D}} \left( 
  \exp\left(-\sum_{i=1}^D\frac{\left(x_i - \frac{1}{3}\right)^2}{k^2}\right) \right. \\ 
  \left. + \exp\left(-\sum_{i=1}^D\frac{\left(x_i - \frac{2}{3}\right)^2}{k^2}\right) 
  \right) ,
\end{multline}
where \( \mathbf{x} = (x_1, x_2,......,x_D) \in \mathbb{R}^D \), with $k = 0.2$.
Note that this function features a strong unbounded multiplicative coupling.

Table.~\ref{tab:camel_dim_scaling} presents the results of our analysis, 
comparing the accuracy of the baseline model against the regularized 
one across different dimensions and dataset sizes.
We observe a consistent improvement in accuracy when moment corrections are applied. 
The extent of this improvement (reported in the fifth column of the table),
strongly depends on both $D$ and $N$: bigger gains can be observed for increasingly large
datasets, but the trend slows down with increasing dimensionality $D$,
as consequence of the so called \textit{curse of dimensionality}.
 
\begin{table}[h]
    \centering
    \begin{ruledtabular}
    \begin{tabular}{ccccc}
        & & \multicolumn{3}{c}{RMSE} \\  
        \cline{3-4}  
        $D$ & $N$ & Without Correction & With Correction & Ratio  \\  
        \hline
        1 & 100    & $1.58 \times 10^{-2}$ & $6.26 \times 10^{-3}$ &  2.52 \\  
        1 & 200    & $8.27 \times 10^{-3}$ & $7.18 \times 10^{-4}$ & 11.51 \\  
        1 & 400    & $4.10 \times 10^{-3}$ & $4.36 \times 10^{-4}$ & 9.40 \\  
        1 & 800    & $2.40 \times 10^{-3}$ & $1.49 \times 10^{-4}$ & 16.10 \\  
        \hline
        3 & 1000   & $4.14 \times 10^{-2}$ & $2.38 \times 10^{-2}$ & 1.74  \\  
        3 & 2000   & $2.82 \times 10^{-2}$ & $1.15 \times 10^{-2}$ & 2.45  \\  
        3 & 4000   & $1.93 \times 10^{-2}$ & $6.76 \times 10^{-3}$ & 2.86  \\  
        3 & 8000   & $1.58 \times 10^{-2}$ & $4.80 \times 10^{-3}$ & 3.29  \\  
        \hline
        6 & 1000   & $8.52 \times 10^{-2}$ & $8.13 \times 10^{-2}$ & 1.05  \\  
        6 & 8000   & $1.68 \times 10^{-2}$ & $1.37 \times 10^{-2}$ & 1.23  \\  
        6 & 50000  & $1.02 \times 10^{-2}$ & $7.46 \times 10^{-3}$ & 1.37  \\  
        6 & 100000 & $7.87 \times 10^{-3}$ & $5.92 \times 10^{-3}$ & 1.33  \\  
    \end{tabular}
    \end{ruledtabular}
    \caption{Convergence of RMSE for $f_{\text{camel}}(\bm{x})$ across different dimensions ($D$) and training sizes ($N$) of random points. 
    The RMSE is computed using 10000 randomly sampled points ($\bm{x} \in [0,1]^D$) outside the training set. The ratio column represents the improvement due to the correction.}
    \label{tab:camel_dim_scaling}
\end{table}

To further investigate the method, we consider another example of a high-dimensional function: the 6D Ackley function, which exhibits weaker coupling between variables compared to the Camel function in the same number of dimensions:
\begin{multline}\label{ackley6d}
   f_{\text{ackley}}(\mathbf{x}) = -20 \exp\left(-0.2 \sqrt{\frac{1}{6} \sum_{i=1}^{6} x_i^2} \right) 
\\- \exp\left( \frac{1}{6} \sum_{i=1}^{6} \cos(2\pi x_i) \right) 
+ 20 + \exp(1), 
\end{multline}
where \( \mathbf{x} = (x_1, x_2,......,x_6) \in \mathbb{R}^6 \). As with the 6D Camel function, we observe that the proposed correction scheme progressively improves the accuracy as the grid becomes denser, as summarized in Table~\ref{tab:ackley_scaling}. This motivates a closer examination of the method's performance in the sparse-data regime, where training points are limited. A detailed analysis of the trade-offs between accuracy and computational cost, alongside with comparisons to other methods, is provided in Appendix~\ref{sec:a3}.

\begin{table}[h]
    \centering
    \begin{ruledtabular}
    \begin{tabular}{ccccc}
        & & \multicolumn{3}{c}{RMSE} \\  
        \cline{3-4}  
        $D$ & $N$ & Without Correction & With Correction & Ratio  \\  
        \hline
        6 & 2000   & $2.91 \times 10^{-2}$ & $2.50 \times 10^{-2}$ & 1.16  \\  
        6 & 4000   & $2.47 \times 10^{-2}$ & $2.01 \times 10^{-2}$ & 1.22  \\  
        6 & 8000   & $2.02 \times 10^{-2}$ & $1.46 \times 10^{-2}$ & 1.38  \\ 
        6 & 16000  & $1.75 \times 10^{-2}$ & $1.10 \times 10^{-2}$ & 1.59  \\
        6 & 32000  & $1.43 \times 10^{-2}$ & $8.68 \times 10^{-3}$ & 1.65  \\
        6 & 64000  & $1.13 \times 10^{-2}$ & $6.27 \times 10^{-3}$ & 1.80  \\
        6 & 100000 & $9.98 \times 10^{-3}$ & $4.27\times 10^{-3}$  & 2.33   \\  
    \end{tabular}
    \end{ruledtabular}
    \caption{Convergence of RMSE for $6D$ $f_{\text{ackley}}(\bm{x})$ across different training sizes ($N$) of random points. 
    The RMSE is computed using 10000 randomly sampled points ($\bm{x} \in [0,1]^6$) outside the training set. The ratio column represents the improvement due to the correction.}
    \label{tab:ackley_scaling}
\end{table}

\begin{figure}[!htb]
    \centering
    \includegraphics[width=0.85\columnwidth]{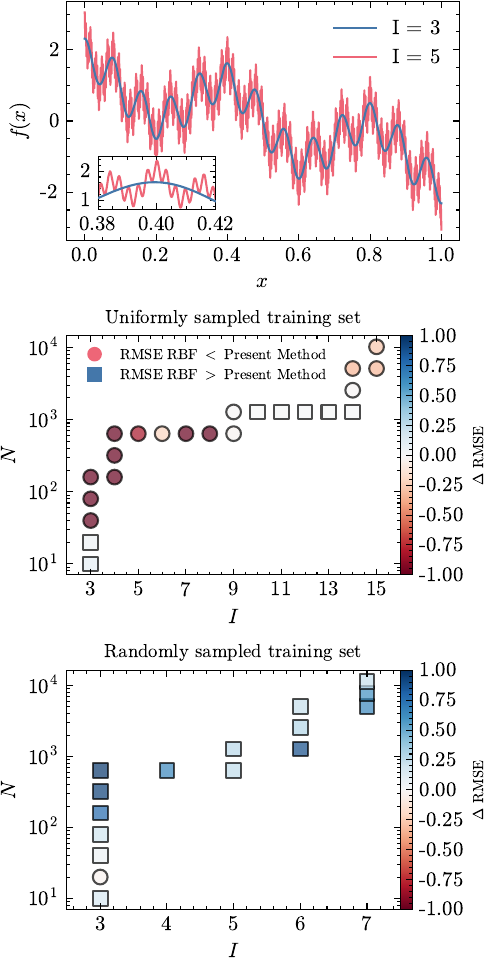}
    \caption{Top panel: 1D Weierstrass function (Eq.~\ref{eq:weier}) with $I = 3$ and $I=5$. 
             Middle and bottom panels: comparison of the error produced by the RBF and present method,
             with colors representing the relative difference between the two methods, 
             defined as $\Delta \mathrm{RMSE} = \left(1 - \frac{\mathrm{RMSE}_{\mathrm{present}}}{\mathrm{RMSE}_{\mathrm{RBF}}}\right)$. 
             In the middle panel the training set is defined using uniform sampling, while the bottom panel uses random sampling.
    }\label{fig:w1d}
\end{figure}

\subsection{Interpolation of a fractal curve}
We now consider the Weierstrass function, a well-known fractal function, 
defined in one dimension. It serves as a classic example of a continuous function 
that is \textit{nowhere differentiable}, making it highly irregular and very 
challenging for interpolation methods; indeed it represents the worst-case scenario 
for interpolators that assume smoothness or differentiability, 
such as polynomial interpolation or spline interpolation.

We use the following definition, consisting of $I$ terms:  
\begin{equation}\label{eq:weier}
  f_{\text{ws}}(x) = \sum_{i=0}^{I-1} \left( \frac{3}{4} \right)^i \cos\left(5^i \pi x\right).
\end{equation}
In the top panel of Fig.~\ref{fig:w1d}, we show two examples 
with $I=3$ (blue curve) and $I=5$ (red curve). The plot highlights
that as $I$ increases, the function exhibits (self-similar) finer-scale oscillations, 
and while remaining continuous everywhere it becomes increasingly rugged and non-smooth, 
ultimately being nowhere differentiable as $I \to \infty$. 
We will hereafter refer to the number of terms $I$ as the complexity of the function. 

To assess performance, we compare our method against a RBF approach, 
varying both the sampling size $N$ and the function complexity $I$.  
For the RBF interpolation, we use a Gaussian kernel and tune 
its free parameter on the test dataset using the same procedure 
applied to the parameter $\theta$ in our method, to ensure a fair comparison.
We begin with a relatively simple case, using $I=3$ terms and a randomly selected sample of $N=20$ points. 
As the sampling density increases, both methods improve in their ability to approximate the function. 
Once both methods achieve satisfactory performance on the test set, we incrementally increase 
the complexity $I$ and track the prediction accuracy using the test RMSE.  
This process of sequential complexity increase and grid refinement is applied 
in two distinct numerical experiments: first, with uniform sampling (Fig.~\ref{fig:w1d}b), 
where training points are spaced regularly along the domain; and second, with random sampling (Fig.~\ref{fig:w1d}c), 
where points are chosen randomly.
It follows that the bottom-left region in Figs.~\ref{fig:w1d}b-c corresponds to a sparse sampling 
grid with low complexity, while the top-right region indicates a denser training set combined
with a higher complexity.

In the figure, the colormap illustrates the relative difference between the two methods, 
defined as $\Delta \mathrm{RMSE} = \left(1 - \frac{\mathrm{RMSE}_{\mathrm{present}}}{\mathrm{RMSE}_{\mathrm{RBF}}}\right)$. 
In particular, red colors (with circle markers) indicate cases where RBF provides a better approximation 
than the present method, while blue colors (with square markers) indicate the opposite case.

We observe that uniform sampling benefits the RBF interpolation, which consistently outperforms 
the present method to varying degrees, particularly in cases with lower complexity and sparser sampling. 
As both the complexity and the number of sampling points increase, the performance gap narrows, 
with the present method progressively catching up to the RBF interpolation.  
However, the trend reverses in the case of random sampling. The plain RBF struggles 
in comparison to the present method, as can be observed in the bottom panel 
of Fig.~\ref{fig:w1d}. The reason comes from the fact that when data is sampled randomly, 
clusters of data points naturally arise, resulting in an uneven distribution across the domain. 
Such clustering leads to ill-conditioned interpolation matrix, since certain regions 
become oversampled while others remain under-sampled, in turn leading to numerical instabilities and inaccuracy.

This issue can be mitigated combining RBF with regularization techniques or by 
preprocessing the data to ensure a more uniform distribution of points. 
While these approaches require careful parameter tuning, 
our method leverages a physics-based regularization inherently,
eliminating the need for extra parameter tuning.

\begin{figure}[htb!]
  \centering
  \includegraphics[width=0.7\columnwidth]{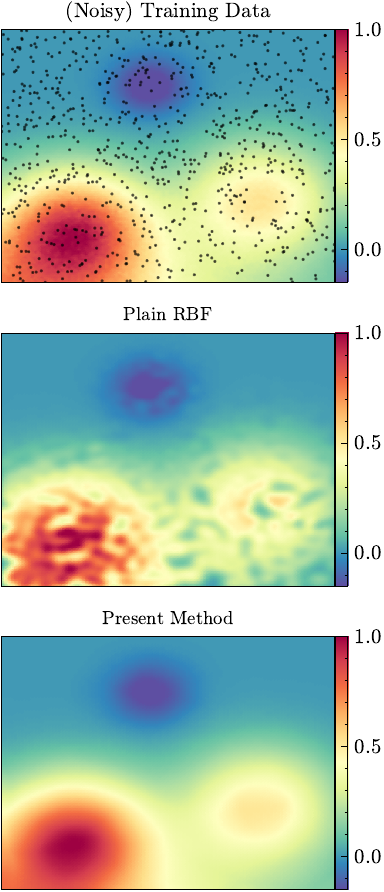}
  \caption{Comparison of 2D Franke function predictions for $5 \%$ 
           Gaussian noise ($s=0.05$) trained and tested on a random set 
           of $1000$ and a uniform set of $10000$ points respectively. 
           Top panel: (noisy) training data. Middle panel: RBF prediction. Bottom panel: prediction with present method. 
  }\label{fig:noisy_franke}
\end{figure}

\subsection{Regression on noisy data}\label{sec:noise}
\begin{figure*}[ht!]
  \centering
  \includegraphics[width=0.85\textwidth]{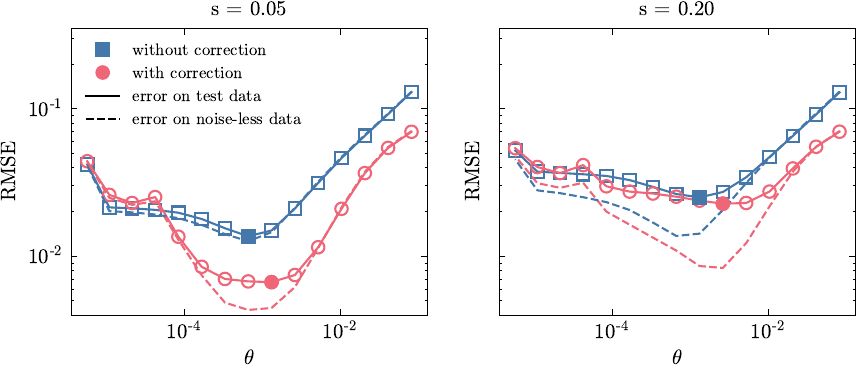}
  \caption{ Energy/error landscape as a function of the temperature parameter $\theta$, for the interpolation 
            of the 2D Franke function with two different noise levels using $1000$ training points: left) $s=0.05$ and right) $s=0.2$.
  }\label{fig:theta_vs_rmse_noise_gaussian_franke2d}
\end{figure*}

In this section, we apply our regularization technique to perform regression on noisy data.  
We revisit the Franke function (Eq.~\ref{eq:franke}) as the base function, this time introducing a noise term:  
\begin{equation}\label{eq:noisy-function}
  f_{\rm noisy}(\bm{x}) = f_{\rm base}(\bm{x}) \left(1 + s \epsilon \right),
\end{equation}  
where $\epsilon$ is a random variable drawn from a Gaussian distribution, and $s$ is a scale factor that controls the noise intensity.  
For example, $s = 0.05$ corresponds to $5\%$ noise in the data. 
The Gaussian noise is added with $\sigma = 1/3$ and $\mu = 0$, based on the $6 \, \sigma$ rule.  
This ensures that most data points fall within the range $[-1, 1]$ ($\mu \pm 3\sigma$), with only a small number of outliers.
Since the focus is on de-noising the data, the goal is to minimize the error relative to the (generally unknown) 
base function on the test set. This represents the hidden minima, which remains unknown to the model, 
while the training is carried out using the noisy data itself.

Following this procedure, we establish a dataset of $N=1000$ randomly sampled points, with 
in Fig.~\ref{fig:noisy_franke}a an example for $s = 0.05$.
Fig.~\ref{fig:noisy_franke}b shows the prediction provided by the same RBF method employed in the previous section.
It can be observed how even a small amount of noise can cause RBF to overfit the noisy data and hence, 
fail to identify the noiseless ground truth.
Fig.~\ref{fig:noisy_franke}c show the results produced by our method; as we can see the physics-based
regularization is highly effective in filtering out the noise, producing results in much better
agreement with the underlying base function (cf. Fig.~\ref{fig:franke_noiseless}).

\begin{figure}[htb]
    \centering
    \includegraphics[width=0.99\columnwidth]{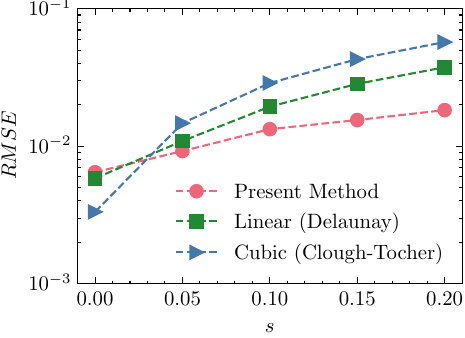}
    \caption{ Interpolation of data from numerical simulations, superimposed with Gaussian noise (eq.~\ref{eq:noisy-function}).
              Comparing the accuracy of the presented method with linear and cubic interpolators for increasingly large
              noise levels.
            }\label{fig:fr_noise_vs_rmse}
\end{figure}

\begin{figure*}[htb]
  \centering
  \includegraphics[width=0.99\textwidth]{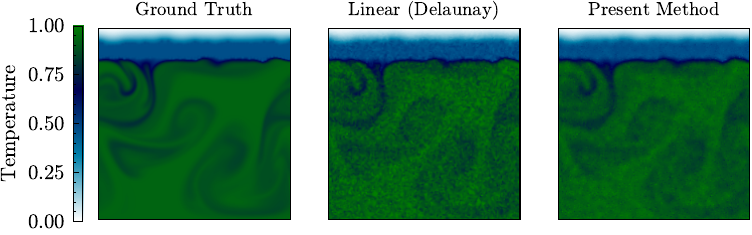}
  \includegraphics[width=0.99\textwidth]{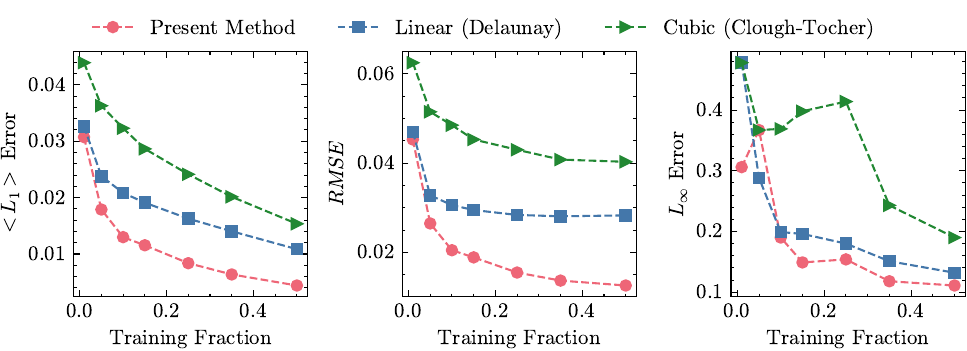}
  \caption{The top three panels show the ground truth, i.e., the noiseless numerical data which we aim to 
           reproduce and the results obtained using Delaunay triangulation (Linear) and the present method respectively;
           here $25\,\%$ of data points were used for the training.
           The bottom panels show the $<L_1>$, $\rm RMSE$, and $L_{\infty}$ error norms, while increasing the amount
           of data used to train the model. We compare the present method with linear and cubic interpolation methods
           with a noise level of $(s=0.15)$. The error metrics are computed using the data not supplied in the training set.
          }\label{fig:fr_tf}
\end{figure*}

\begin{figure*}[htb]
  \centering
  \includegraphics[width=0.99\textwidth]{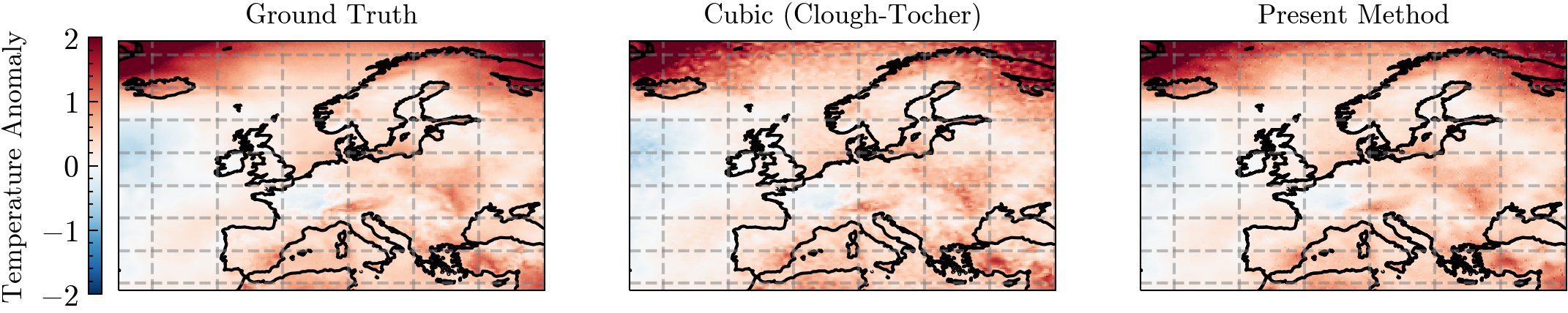}
  \includegraphics[width=0.99\textwidth]{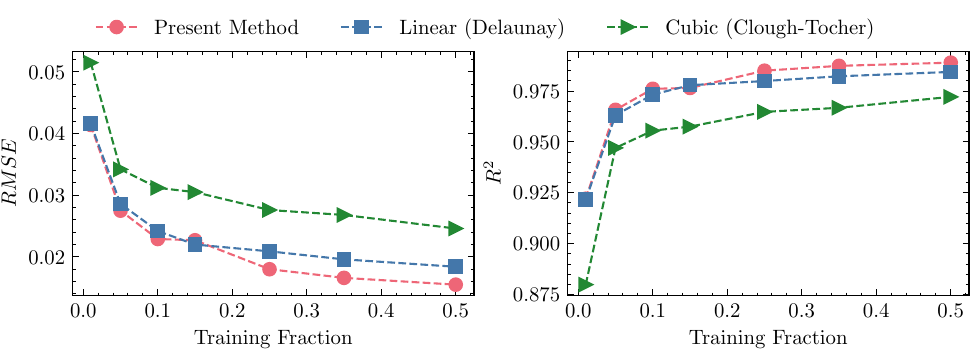}
  \caption{The top three panels show the ground truth, i.e., the noiseless recorded temperature anomaly             data (from Copernicus Climate Change Service) which we aim to 
           reproduce and the results obtained using Clough-Tocher triangulation (Cubic) and the present method respectively;
           here $25\,\%$ of data points were used for the training.
           The bottom panels show the $\rm{RMSE}$ and the coefficient of determination $R^2$ with respect to the ground truth, while increasing the amount
           of data used to train the model. We compare the present method with linear and cubic interpolation methods
           with a noise level of $(s=0.35)$. The error metrics are computed using the data not supplied in the training set.
          }\label{fig:weather_tf}
\end{figure*}

The effect of regularization is more clearly illustrated in Fig.~\ref{fig:theta_vs_rmse_noise_gaussian_franke2d}, 
where we investigate the impact of correction terms on the overall accuracy.  
The plot in the left panel shows the energy landscape as a function 
of the temperature parameter $\theta$ for $s = 0.05$.  
The continuous lines represent the error on the noisy test data, 
while the dotted lines show the error with respect to the base (noise-less) data.
When moment-based corrections are applied, we observe a significant reduction 
in the error on the noisy test data compared to the case without corrections.  
Additionally, the corrections enable the model to reach a better local minimum 
on the error curve for the noise-less data, thus improving the overall prediction accuracy.

As the signal-to-noise ratio decreases, the advantage of the moment correction weakens.  
This is shown in the right panel of Fig.~\ref{fig:theta_vs_rmse_noise_gaussian_franke2d}, 
where the same analysis is repeated for $s = 0.2$. On the positive side, while the moment 
corrections become less effective as noise increases, they still perform at least as well as the 
baseline method without corrections, ensuring that the model does not worsen compared to the non-regularized approach.

Next, we consider a dataset derived from numerical simulations of convective fluid flows, 
where the underlying analytical form is unknown. This case study illustrates a typical example of 
application, where a smaller, finite set of data points is used to generate data at a finer scale. 
A similar situation arises when dealing with experimental data, which is often sparse and noisy. 
In such cases, interpolation methods need to be both fast and robust, 
particularly when the aim is to generate more accurate, noise-free data from a limited dataset.

For this study, we use a $240 \times 240$ dataset from a snapshot of a 2D temperature field 
obtained from numerical simulations of phase-change convection  (see Ref.~\cite{huber2008lattice} for details on the numerical setup).
For model training, $25\%$ of the dataset is randomly selected 
and split into training and validation sets in an $80:20$ ratio. 
In other words, one-fourth of the data is used for training, while the remaining 
three-fourths are reserved for testing the model’s predictions. 
To mimic a noisy dataset, we use Eq.~\ref{eq:noisy-function} to introduce varying degrees of Gaussian noise~\cite{weiJCP,SavgolOptimum}. 
We compare our method with two commonly used interpolation techniques: linear and cubic interpolation. 
For linear interpolation, we use Delaunay triangulation~\cite{ChenDelaunay}, 
where the value at an unknown point is estimated by performing piecewise linear interpolation 
based on the nearest known points that form a triangle. 
For cubic interpolation, we employ the Clough-Tocher method~\cite{cloughtoucher}, 
which fits cubic polynomials ($C^1 \text{ continuous}$) to the triangles, 
resulting in a smoother interpolation compared to the linear method, 
which only ensures $C^0 \text{ continuity}$ (i.e., piecewise linear).

In Fig.~\ref{fig:fr_noise_vs_rmse}, we compare the RMSE error on the validation set 
for different levels of noise in the data. 
As expected, interpolation methods provide more accurate predictions when working 
on noise-less data, with higher-order methods yielding the best accuracy. 
However, as the noise level increases, the present method consistently outperforms 
both interpolation schemes. Note that lower-order methods can be seen as a form of regularization; 
in this case, the linear interpolation method proves more robust than 
the cubic interpolator, which is more susceptible to overfitting the noisy data.

Considering now the case with $s = 0.15$, for which top panel in Fig.~\ref{fig:fr_tf} offers a visual representation,
we evaluate the effect of increasing the number of points in the training set. 
The bottom panel shows the trend for three different error metrics: $<L_1>$, RMSE, and $L_{\infty}$, 
as the fraction of training points increases. The present method outperforms the interpolation schemes across all metrics. 
Notably, the RMSE analysis reveals that both the linear and cubic interpolation methods reach a plateau already 
when only $10\%$ of the data is used, meaning adding more data does not improve their performance. 
In contrast, the present method continues to improve well beyond that point, allowing a more effective use of
additional data.\\
As a final example, more closely aligned with real-world applications involving experimental data, we consider observational data from the Climate Data Store (CDS) of the Copernicus Climate Change Service (C3S)~\cite{C3S_CDS}, focusing on the European region. The dataset comprises 153 latitude points and 261 longitude points as spatial coordinates. The ground truth, shown in the first figure of the top panel in Fig.~\ref{fig:weather_tf}, represents the near-surface (approximately 2m) temperature anomaly for the year 2016. This anomaly is computed as the difference between the pointwise temperature in 2016 and the long-term average temperature from the 1991-2020 climate reference period.
We inject varying degrees of Gaussian noise to the data~\cite{ZurNoiseInjection2009, CHEN2022916}. Figure~\ref{fig:weather_tf} illustrates the case with $35\%$ noise, higher than in the previous example, to explore the method’s robustness under increased uncertainty. As seen in the second and third plots of the top panel, the proposed method yields a visibly smoother and more accurate reconstruction compared to the Clough-Tocher approach. This improvement is further supported by the bottom panel of Fig.~\ref{fig:weather_tf}, which shows lower RMSE and higher $R^2$ scores across increasing training fractions, clearly outperforming both Delaunay and Clough-Tocher methods.

\subsection{Computational Cost}

\begin{figure}[htb]
  \centering
  \includegraphics[width=0.99\columnwidth]{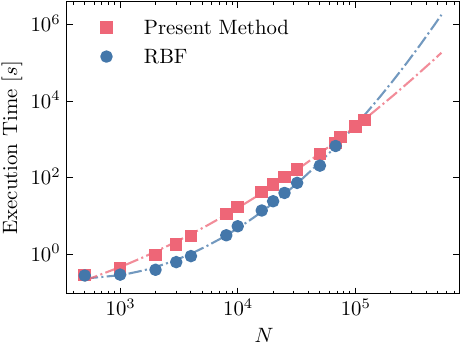}
  \caption{   
    The representative computational cost of the Lagrange multiplier calculations 
    for each training point in the present method is compared with the weight 
    determination step of the Radial Basis Function (RBF) across a range of datasets. 
    It is evident that for asymptotically large $N$ (around $10^5$), 
    the present method exhibits better efficiency.
  }\label{fig:time_complexity}
\end{figure}

In this section, we analyze the computational costs of the present method in terms of storage requirements and speed. 
We consider the RBF method as reference, given the underlying similarities in the two approaches.

The most computationally intensive step in Algorithm~\ref{alg:summary} is the calculation of first-order corrections, 
which involves solving for the Lagrange multipliers (LM). 
This step requires solving $D$ simultaneous equations at each training point to compute the Lagrange multipliers. 
In contrast, the RBF method requires solving a set of $N$ coupled algebraic equations (cf. Eq.~\ref{eq:phi-rbf}) 
to compute a set of unknown weights.

As a result, the RBF method stores $N_t$ points to hold the weights, while the present method 
stores $N \times D$ values for the Lagrange multipliers and $N_t$ self-prediction values 
for the second-moment correction. Additionally, the global nature of the RBF method requires temporary storage 
of an $N^2$ matrix during training, making it significantly more memory-intensive than the pointwise calculations 
in the present method, which only requires $D^2$ matrices.

We compare the average time required to compute the validation error for each $\theta_i$ generated using the 
search scheme described in Sec.~\ref{sec:theta}, which provides a basis for estimating the training time of both methods.

In order to establish a fair comparison, we use NumPy’s $linalg$ package to solve the linear equations 
involved in both the Lagrange multiplier and weight calculations. We perform tests using training sets of
increasing sizes, recording computation times for both methods.

As shown in Fig.~\ref{fig:time_complexity}, the RBF method is initially faster than the present method. 
However, the time required for RBF increases more rapidly with $N$, 
as solving $\mathcal{O}(N)$ simultaneous equations becomes increasingly costly,
with a crossover occurring around $N \sim \mathcal{O}(10^5)$, beyond which the present method is more efficient.
The absence of data points for RBF beyond $N = 10^5$ in Fig.~\ref{fig:time_complexity} 
is due to the fact that the system matrix becomes too large to be stored on the machine used for the test,
whereas no such limitation is encountered with the present method.

\section{Conclusions}\label{sec:conclusion}

In conclusion, we have introduced a novel physics-based regularization technique 
for function learning, inspired by statistical mechanics. 
By drawing an analogy between the task of interpolating a dataset and minimizing the energy of a system, 
we have introduced physics-inspired corrections that significantly enhance accuracy.

We have demonstrated that by constraining the lower-order moments of the (data) distribution, 
we can minimize the discrepancy between its discrete and continuum counterparts. 
This approach leads to the discovery of favorable local minima in the energy landscape, 
resulting in improved performance for both interpolation and regression tasks, even in high-dimensional spaces.

Notably, and unlike traditional regularization techniques, our method does not require 
the empirical tuning of free parameters, a feature especially advantageous 
for regression tasks involving noisy data.

Moreover, we have shown that the local nature of the method provides computational advantages,
both in terms of cost and memory requirements, compared to Radial Basis Function interpolators, 
particularly for large-scale datasets.

While the primary focus of our numerical analysis has been highlighting the flexibility and robustness 
of our method, with no claims of superiority over other approaches, in
future works we plan to delve deeper into the method's potential applications. 
For example, we plan to explore its use as a preprocessing step in deep learning pipelines.
Another interesting prospect is the applicability of the method for the definition of
smoothing kernels and grid-free formulations for computational fluid dynamics solvers~\cite{sph-kernels,Strzelczyk2024}.

\section*{Data availability statement} 

An implementation of the algorithm, alongside scripts and datasets to reproduce
the results presented in this work, can be found in the following GitHub repository:
\href{https://github.com/agabbana/kinetic-based-regularization}{https://github.com/agabbana/kinetic-based-regularization}

\section*{Acknowledgments} 
We would like to acknowledge Prof. M. Vidyasagar and
Andrea Montessori for insightful discussions. 
S. A. acknowledges support from the Abdul Kalam Technology Innovation National Fellowship (INAE/SA/4784).
A.G. gratefully acknowledges the support of the U.S. Department of Energy through the 
LANL/LDRD Program under project number 20240740PRD1 and the Center for Non-Linear Studies for this work.

\appendix
\section{Calculation of Lagrange Multipliers}\label{sec:a1}
\begin{figure*}[htb]
    \centering
    \includegraphics[width=0.99\textwidth]{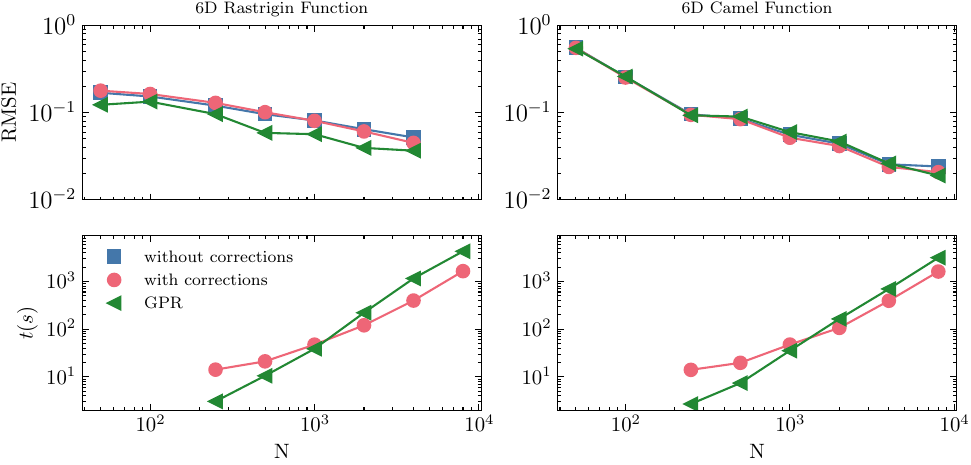}
    \caption{Performance comparison between Gaussian Process Regression (GPR) and the present method in the $6D$ sparse-grid regime, for the Rastrigin function (uncoupled) and Camel function (multiplicatively coupled). Top panel shows the RMSE on a random test set of 10,000 points with varying training set sizes $N$. Bottom panel shows the corresponding computation times.}
    \label{fig:sparse_grid}
\end{figure*}

We are interested in finding an interpolated value at a point $\bm{x}$. 
In the expression 
$\hat{\phi}(\bm{x}) = \sum_i \phi(\bm{x_i}) P(\|\tilde{\bm{\xi}}_i||, \theta) = \sum_i \phi(\bm{x_i}) P(\|\tilde{\bm{x}} - \bm{x_i}\|, \theta) $, 
we inherently assume a zero correction, i.e., $\tilde{x} = x$. 
However, we aim to construct a distribution function that satisfies the first moment condition given in Eq.~\eqref{eq:first_moment}. 
For a locally linear approximation of a function $\phi(x)$ in 1D, we impose the condition
\begin{equation}\label{eq:root_simple}
    \begin{split}
        \sum_i (x - x_i) P(\|\tilde{x}^{(k+1)}-x_i\|,\theta) = 0,
    \end{split}
\end{equation}
where $k+1$ denotes the iteration index (to be discussed below).

Writing $\tilde{x} = x + \delta x$ and applying an iterative procedure, we have 
$\tilde{x}^{(k+1)}=\tilde{x}^{(k)}+\delta\tilde{x}^{(k)}$, where $\tilde{x}^{(0)}=x$. 
It can be shown that $P$ in Eq.~\eqref{eq:root_simple} does not require a normalization, allowing us to work with the non-normalized Gaussian distribution:
\begin{gather}
        P(\|\tilde{x}-x_i\|,\theta) = \exp\left(-\frac{(\tilde{x} - x_i)^2}{2\theta}\right).
\end{gather}

At iteration $k$, the condition reads
\begin{equation}
    \sum_i (x - x_i) \exp\left(-\frac{(\tilde{x}^{(k)} + \delta \tilde{x}^{(k)} - x_i )^2}{2\theta}\right) = 0.
\end{equation}

Neglecting terms of order $(\delta x)^2$ and omitting the superscript $(k)$, we obtain a transcendental equation for $\delta x$:
\begin{equation}
    \sum_i (x - x_i) \exp\left(-\frac{(\tilde{x}-x_i )^2}{2\theta}\right)\exp\left(-\frac{\delta \tilde{x}(\tilde{x}  - x_i )}{\theta}\right) = 0.
\end{equation}

Linearizing assuming small $\delta x$,
\begin{gather}
    \sum_i (x - x_i) \exp\left(-\frac{(\tilde{x}  - x_i )^2}{2\theta}\right)\left(1-\frac{\delta \tilde{x}(\tilde{x}  - x_i )}{\theta}\right) = 0. \nonumber \\[1em]   
    \sum_i (x_i - x)P(\|\tilde{x}-x_i\|,\theta) \left(1-\frac{1}{\theta}\delta \tilde{x}(\tilde{x}  - x_i )\right) = 0 . \label{eq:exp_gaussian}
\end{gather}

Expanding and collecting terms,
\begin{multline}
        \sum_i (x - x_i) P(\|\tilde{x}-x_i\|,\theta) \\ 
        -  \delta \tilde{x} \sum_i  \frac{1}{\theta}(\tilde{x}-x_i )(x-x_i) 
   P(\|\tilde{x}-x_i\|,\theta)  = 0 ,
\end{multline}
which yields the solution
\begin{gather}
    \delta \tilde{x} = \theta \frac{\sum_i (x - x_i) P(\|\tilde{x}-x_i\|,\theta)}{\sum_i\left((\tilde{x}-x_i )(x-x_i) P(\|\tilde{x}-x_i\|,\theta)\right)}.      
\end{gather}

This iteratively becomes,
\begin{gather}
    \delta \tilde{x}^{(k)} = \theta \frac{\sum_i (x - x_i) P(\|\tilde{x}^{(k)}-x_i\|,\theta)}{\sum_i\left((\tilde{x}^{(k)}-x_i )(x-x_i) P(\|\tilde{x}^{(k)}-x_i\|,\theta)\right)},
\end{gather}
and hence 
\begin{gather}
    \tilde{x}^{(k+1)} = \tilde{x}^{(k)}+\delta \tilde{x}^{(k)} ,\\ \nonumber
    \tilde{x}^{(k+1)} = \tilde{x}^{(k)} + \theta \frac{\sum_i (x - x_i) P(\|\tilde{x}^{(k)}-x_i\|,\theta)}{\sum_i\left((\tilde{x}^{(k)}-x_i )(x-x_i) P(\|\tilde{x}^{(k)}-x_i\|,\theta)\right)} .
\end{gather}

Once we have the corrected $x$, we use this value for the calculation of the values of the distribution function. 
Extending this to $D$ dimensions where $\bm{x} = x_\alpha \hat{e}_\alpha$, the algebra in Eq.~\ref{eq:exp_gaussian} generalizes to
\begin{multline} \label{eq:LM_general}
    \sum_i (x_{\alpha} - x_{\alpha,i}) P(\|\tilde{\bm{x}}-\bm{x_i}\|,\theta)\\
    - \frac{\delta \tilde{x}_{\kappa}}{\theta} \sum_i 
    (\tilde{x}_{\kappa} - x_{\kappa,i})  
    (x_{\alpha} - x_{\alpha,i}) P(\|\tilde{\bm{x}}-\bm{x_i}\|,\theta) = 0, 
\end{multline}
where, $\kappa$ is a repeated index, and the above represents a linear algebra problem.

Letting $P(\|\tilde{\bm{x}} - \bm{x}_i\|, \theta) = P_i$ for brevity, in 2D, representing coordinates as $\bm{x}_i = x_i \hat{e}_x + y_i \hat{e}_y$, 
Eq.~\eqref{eq:LM_general} translates to the matrix problem $\mathcal{A} : \delta \bm{X} = \bm{B}$, where
\begin{gather}
    \mathcal{A} = \frac{1}{\theta}
    \begin{bmatrix}
        \sum_i (x - x_i)(\tilde{x} - x_i)P_{i} & \sum_i (x - x_i)(\tilde{y} - y_i)P_{i} \\ 
        \sum_i (\tilde{x} - x_i)(y - y_i)P_{i} & \sum_i (y - y_i)(\tilde{y} - y_i)P_{i}
    \end{bmatrix}, \nonumber \\ 
    \pmb{B} = 
    \begin{bmatrix}
        \delta \tilde{x} \\ \delta \tilde{y}
    \end{bmatrix}, \nonumber\\ 
    \delta \bm{X} = 
    \begin{bmatrix}
        \sum_i (x - x_i)P_{i} \\ \sum_i (y - y_i)P_{i}
    \end{bmatrix}.
\end{gather}

This system can be solved iteratively for each point $\bm{x}$ to obtain the correction $\delta \bm{X}$.

\section{Hermite polynomial expansion approach}\label{sec:a2}
\begin{figure*}[htb]
    \centering
    \includegraphics[width=0.99\textwidth]{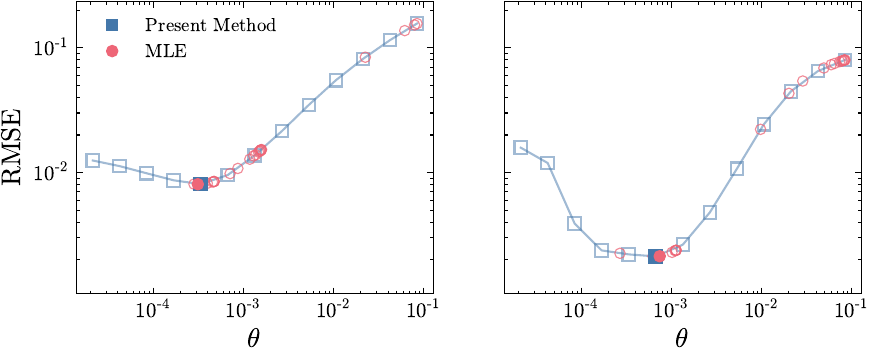}
    \caption{Optimal temperature estimation for the 2D Franke function using 2000 random sample points ($N = 2000$). The left/right panel shows results without/with corrections. Filled markers indicate the minimum values. In both cases, the present method converges in 13 iterations, while MLE requires 46 and 52 iterations, respectively.}
    \label{fig:mle}
\end{figure*}
Let us consider a target function, \(\phi(\bm{x})\), which admits a second-order Hermite expansion~\cite{grad1949hermite}:
\begin{equation}
    \phi(\bm{x}) = a_0\, \mathcal{H}^{(0)}(\bm{x}) 
    + a_{1\,\alpha}\, \mathcal{H}^{(1)}_\alpha(\bm{x}) 
    + a_{2\,\alpha\beta}\, \mathcal{H}^{(2)}_{\alpha\beta}(\bm{x}) + \cdots.
\end{equation}
Here, $\alpha$ and $\beta$ are representatives of dimensional indices. The coefficients can be computed via projection with respect to a Gaussian weight:
\begin{gather}
a_{0}(\bm{x}) = \int \exp\left(-\frac{\bm x ^2}{2\theta}\right)\, \phi(\bm{x})\, d\bm{x}, \\ \nonumber
a_{1\,\alpha}(\bm{x}) = \int x_\alpha \, \exp\left(-\frac{\bm x ^2}{2\theta}\right)\, \phi(\bm{x})\, d\bm{x}, \\ \nonumber
a_{2\,\alpha\beta}(\bm{x}) = \int( x_\alpha \, x_\beta  - \theta\delta_{\alpha\beta} ) 
\exp\left(-\frac{\bm x ^2}{2\theta}\right)\, \phi(\bm{x})\, d\bm{x}.
\end{gather}

Note that as the third-order contribution is ignored, errors will be proportional to the third Hermite coefficient.  
Then for a given point of interest ${\bm x_i}$, our approximate projection is given by,
\begin{equation}
    \hat \phi (\bm x_i) = \int \phi(\bm{x}) P(\bm \xi_i ,\theta) \, d \bm{x},
\end{equation}
where $\bm \xi_i = ||\bm x- \bm x_i|| $. Substituting the form of $\phi(\bm{x})$, we have
\begin{align}
\hat \phi({\bf x_i}) = \int \big(& a_0\, {\cal H}^{(0)}({\bm x}) 
+ a_{1\, \alpha } \,{\cal H}^{(1)}_{\alpha }({\bm x}) \nonumber \\
& + a_{2\, \alpha \beta } \,{\cal H}^{(2)}_{\alpha \beta }({\bm x}) 
+ \cdots \big) P(\bm \xi_i ,\theta) \, d \bm{x}
\end{align}

Here, the constants are not a function of $\bm x$  and can be taken out of the integral. Therefore we have,
\begin{multline}
\tilde \phi({\bm x_i})= a_0 \int  {\cal H}^{(0)}({\bm x}) P(\bm \xi_i ,\theta) d \bm{x} \\
+a_{1\, \alpha } \int {\cal H}^{(1)}_{\alpha }({\bm x})  P(\bm \xi_i ,\theta) \,d \bm{x}\\ 
+a_{2\, \alpha \beta } \int \,{\cal H}^{(2)}_{\alpha \beta }({\bm x})  P(\bm \xi_i ,\theta)\,d \bm{x}  +\cdots 
\end{multline}

Notice that the orthogonal Hermite polynomials are
${\cal H}^{(0)} = 1$, ${\cal H}^{(1)} = x_\alpha$, ${\cal H}^{(2)} = x_{\alpha}x_{\beta} - \theta \delta_{\alpha \beta}$ etc. Substituting these forms in the equation above, we get
\begin{multline}
\hat \phi({\bm x_i})= a_0 \int  P(\bm \xi_i ,\theta) \, d \bm{x} + a_{1\, \alpha } \int x_{\alpha}  P(\bm \xi_i ,\theta) \, d \bm{x} \\ 
+ a_{2\, \alpha \beta } \int (x_{\alpha} x_{\beta} - \theta \delta_{\alpha\beta})  P(\bm \xi_i ,\theta) \, d \bm{x} + \cdots
\end{multline}

From here, we can see that if the first-order correction, that is, the Lagrange multiplier $\tilde {\bm {\xi}_i}$ in place of the naive $\bm \xi_i$ (or the point of interpolation $\bm x_i$) is used, we get.
\begin{align}
\phi({\bm x_i}) = a_0 + a_{1\, \alpha} x_{i\,\alpha}
+ a_{2\, \alpha \beta} (x_{i\,\alpha} x_{i\,\beta} + \theta \delta_{\alpha \beta}) \notag \\
- a_{2\, \alpha \beta} \int \theta \delta_{\alpha \beta} P(\bm {\tilde {\xi}}_i, \theta) d\bm x \notag \\
\implies \phi({\bm x_i}) = a_0 + a_{1\, \alpha} x_{i\,\alpha}
+ a_{2\, \alpha \beta} x_{i\,\alpha} x_{i\,\beta} + \cdots
\end{align}

Hence, we get the exact Hermite representation up to the first order here. It is easy to see that if the second-order correction, i.e, the self-correction is implemented as well, we retrieve the exact Hermite representation up to second order, namely
\begin{gather}
 \phi({\bm x_i})= a_0  +a_{1\, \alpha } x_{i\,\alpha}
+a_{2\, \alpha \beta }  \left(\, x_{i\,\alpha} x_{i\,\beta} -\theta\delta_{\alpha\beta}\right) +\cdots \notag \\
\text{or, }\phi({\bm x_i})= a_0 \mathcal{H}^{(0)}  +a_{1\, \alpha } \mathcal{H}^{(1)}
+a_{2\, \alpha \beta }  \mathcal{H}^{(2)} +\cdots .
\end{gather}
Hence we see that the proposed corrections work in the Hermite space as well, recovering the expansion up to second-order at the interpolation points. The present formulation can be interpreted as the discrete limit of the analysis presented above.

\section{Accuracy and Performance for Sparse Data}\label{sec:a3}
\begin{table*}[htbp]
    \centering
    \begin{ruledtabular}
    \begin{tabular}{cccccccc}
        $N$ & 
        \makecell{No \\ Corrections} & 
        \makecell{No  \\Corrections (MLE)} & 
        \makecell{All \\ Corrections} & 
        \makecell{All \\ Corrections (MLE)} & 
        $t_{\text{MLE}} / t_{\text{present}}$ & 
        $\theta/d^2_{\text{typ}}$ & 
        $\theta/d^2_{\text{typ}}$ (MLE) \\
        \hline
        100  & $4.89 \times 10^{-2}$ & $4.85 \times 10^{-2}$ & $3.38 \times 10^{-2}$ & $3.37 \times 10^{-2}$ & 2.83  & 0.65  & 0.73 \\
        500  & $2.35 \times 10^{-2}$ & $2.33 \times 10^{-2}$ & $1.08 \times 10^{-3}$ & $1.07 \times 10^{-2}$ & 14.06 & 0.69  & 0.76 \\
        1000 & $1.31 \times 10^{-2}$ & $1.31 \times 10^{-2}$ & $4.07 \times 10^{-3}$ & $4.07 \times 10^{-3}$ & 9.75  & 0.73 & 0.76 \\
        2000 & $8.10 \times 10^{-3}$ & $8.07 \times 10^{-3}$ & $2.13 \times 10^{-3}$ & $2.14 \times 10^{-3}$ & 11.05 & 1.54  & 1.43 \\
        4000 & $6.41 \times 10^{-3}$ & $6.41 \times 10^{-3}$ & $8.97 \times 10^{-4}$ & $8.98 \times 10^{-4}$ & 8.42  & 1.50  & 1.52 \\
        8000 & $4.41 \times 10^{-3}$ & $4.37 \times 10^{-3}$ & $6.81 \times 10^{-4}$ & $6.68 \times 10^{-4}$ & 11.38 & 3.05 & 2.24 \\
        12000 & $3.47 \times 10^{-3}$ & $3.45 \times 10^{-3}$ & $4.00 \times 10^{-4}$ & $3.99 \times 10^{-4}$ & 22.62 & 1.15 & 1.41 \\
    \end{tabular}
    \end{ruledtabular}
    \caption{Comparison of hyperparameter optimization ($\theta_{\rm optimal}$) between the present method and Maximum Likelihood Estimation with Conjugate Gradient (MLE-CG), based on validation RMSE and training time. Results are obtained using randomly sampled data from the 2D Franke function.}
    \label{tab:mle_training}
\end{table*}

To assess performance in the sparse data regime, particularly for higher-dimensional problems, we analyze the case where the training set size is small ($N < \mathcal{O}(10^3)$). As a benchmark, we compare our approach with the well-established Additive Kernel Gaussian Process Regression (GPR) method~\cite{duvenaud2011additivegaussianprocesses}, which is particularly effective for high-dimensional uncoupled functions with sparse training data.

We consider the 6D Rastrigin function, a fully separable benchmark function defined as:
\begin{equation}
    f_{\text{ras}}(\mathbf{x}) = \sum_{i=1}^{6} \left[ x_i^2 - A \cos(2\pi x_i) + A \right],
\end{equation}
where $  \mathbf{x} = (x_1, x_2,......,x_6) \in \mathbb{R}^6  $, with $A = 10$.

For GPR we use the built-in Python library $\textit{gaussian process}$. The kernel is constructed as a sum of six Gaussian kernels, one for each dimension:
\begin{equation}
    P(||\mathbf{\xi}||,l) = \sum_{i = 1}^D \text{exp}\left(-\frac{\xi_{i}}{2l_i^2}\right).
\end{equation}
Here, each index $i$ corresponds to a spatial dimension, and the kernel is trained by optimizing the individual length scales $l_i$ for each Gaussian component. The bounds on the length scales are set between $1\%$ of the distance between two neighboring points and 10 times the domain length, ensuring a reasonable training time.
Due to the sparsity and randomness of the sampling grids, the choice of training dataset can strongly affect the results. To mitigate this, all results reported in this study are averaged over 10 independent trials, each using different randomly sampled training sets. The average computation time over three trails is shown in the bottom panel of Fig.~\ref{fig:sparse_grid}.
As shown in the top panel of Fig.~\ref{fig:sparse_grid}, the present method achieves convergence on par with the additive GPR for both functions. 
However, the bottom panel highlights a key advantage of our method: as the number of training points increases, the computational cost of Additive GPR grows significantly, by a factor of 2 to 3, relative to our approach, making our method more scalable for denser grids. We remark that GPR itself is an area of active research, with recent studies~\cite{Manzhos2023}  showing that it is often possible to bypass the need of invoking MLE to determine the optimal kernel parameters, leading to potential additional speedups.

\section{Comparing methods for determining the optimal temperature}\label{sec:a4}

In this appendix section we compare the method for determining the optimal value for $\theta$ described in Sec.~\ref{sec:theta} against a standard approach such as the Maximum Likelihood Estimate (MLE).

We employ MLE to minimize the validation RMSE which is a function of temperature. 
In order to ensure the positivity of $\theta$, we consider the following formulation:
\begin{equation}\label{eq:mle_grad}
    \frac{\partial (RMSE)}{\partial z} = \frac{\partial}{\partial z}\, \left(\frac{\;
                      \Bigl\langle\bigl(\hat{\phi}(\bm{x})-\phi(\bm{x})\bigr)^2\Bigr\rangle^{\tfrac{1}{2}}}{\max|\phi(\bm{x})|}\right),
\end{equation}
where $z = \log(\theta)$. We solve the minimization problem using the Conjugate Gradient (CG) method.

We assess both methods in terms of training time and final validation accuracy. For a fair comparison, the CG optimizer is initialized using the same initial guess from Eq.~\ref{eq:theta-guess}. %

Figure~\ref{fig:mle} shows the error landscape as a function of $\theta$, showing that both methods, whether including corrections or not, converge to similar minima. However, in this example the MLE-CG optimization requires more iterations. %

Detailed training results are summarized in Table~\ref{tab:mle_training}. To quantify computational cost, we define the time ratio
\begin{equation}
\tau = \frac{t_{\rm MLE}}{t_{\rm present}},
\end{equation}
which compares the time taken by MLE to that of the present method (with both first- and second-order corrections). We observe that MLE can be up to an order of magnitude more expensive.

The values of $\theta/d^2_{\rm typ}$ in the table indicate the effective kernel width relative to the local scale set by the $k$-nearest neighbor distance ($k=5$). It is worth remarking that the proposed method builds a list of candidate $\theta$ values based solely on domain coverage (see Eq.~\ref{eq:theta_lst}), independent of validation error. Despite this, the validation RMSE achieved is comparable to that of MLE across all tested datasets.

\begin{table}[htb]
    \centering
    \begin{ruledtabular}
    \begin{tabular}{ccccc}
        \multicolumn{1}{c}{$N$} & 
        \multicolumn{2}{c}{Without Correction} & 
        \multicolumn{2}{c}{With Correction} \\
        \cline{2-3} \cline{4-5}
        & Present & MLE & Present & MLE  \\
        \hline
        100  & $6.16 \times 10^{-2}$ & $6.38 \times 10^{-2}$ & $4.97 \times 10^{-2}$ & $5.08 \times 10^{-2}$ \\
        500  & $2.21 \times 10^{-2}$ & $2.19 \times 10^{-2}$ & $9.06 \times 10^{-3}$ & $8.99 \times 10^{-3}$ \\
        1000 & $1.46 \times 10^{-2}$ & $1.47 \times 10^{-2}$ & $4.25 \times 10^{-3}$ & $4.23 \times 10^{-3}$ \\
        2000 & $9.11 \times 10^{-3}$ & $9.06 \times 10^{-3}$ & $2.69 \times 10^{-3}$ & $2.68 \times 10^{-3}$ \\
        4000 & $6.54 \times 10^{-3}$ & $6.54 \times 10^{-3}$ & $1.28 \times 10^{-3}$ & $1.28 \times 10^{-3}$ \\
        8000 & $4.44 \times 10^{-3}$ & $4.44 \times 10^{-3}$ & $1.00 \times 10^{-3}$ & $9.90 \times 10^{-4}$ \\
        12000 & $3.57 \times 10^{-3}$ & $3.59 \times 10^{-3}$ & $7.91 \times 10^{-4}$ & $7.55 \times 10^{-4}$ \\

    \end{tabular}
    \end{ruledtabular}
    \caption{Comparison of RMSE on deployment data between the present method and MLE-CG for randomly sampled 2D Franke function. Values are reported both with and without corrections.}
    \label{tab:mle_deployment}
\end{table}

For completeness, both methods are evaluated on an independent test set of $10000$ uniformly spread points. Predictions obtained using the respective optimal temperatures (from both methods) are reported in Table~\ref{tab:mle_deployment}. The test performance confirms that, even without relying on validation error, the present method yields nearly equivalent accuracy at a fraction of the computational cost.

\end{document}